\documentclass[letterpaper]{article} 
\usepackage{aaai2026}  
\usepackage{times}  
\usepackage{helvet}  
\usepackage{courier}  
\usepackage[hyphens]{url}  
\usepackage{graphicx} 
\usepackage{booktabs}
\usepackage{multirow}
\usepackage{multicol}
\usepackage{amsmath}
\usepackage{xcolor}

\usepackage{amsfonts}
\usepackage{natbib}  
\usepackage{caption} 
\usepackage{algorithm}
\usepackage{algorithmic}

\usepackage{newfloat}
\usepackage{listings}
\DeclareCaptionStyle{ruled}{labelfont=normalfont,labelsep=colon,strut=off} 
\floatstyle{ruled}
\newfloat{listing}{tb}{lst}{}
\floatname{listing}{Listing}
\title{FilmWeaver: Weaving Consistent Multi-Shot Videos with Cache-Guided Autoregressive Diffusion}
\author{
    Xiangyang Luo\textsuperscript{\rm 1,2}\thanks{This work was conducted during the author's internship at Kling Team, Kuaishou Technology.},
    Qingyu Li\textsuperscript{\rm 2}\thanks{Corresponding authors.},
    Xiaokun Liu\textsuperscript{\rm 2},
    Wenyu Qin\textsuperscript{\rm 2},
    Miao Yang\textsuperscript{\rm 2},
    Meng Wang\textsuperscript{\rm 2},\\
    Pengfei Wan\textsuperscript{\rm 2},
    Di Zhang\textsuperscript{\rm 2},
    Kun Gai\textsuperscript{\rm 2},
    Shao-Lun Huang\textsuperscript{\rm 1$\dagger$}
}
\affiliations{
    \textsuperscript{\rm 1}Tsinghua Shenzhen International Graduate School, Tsinghua University\\
    \textsuperscript{\rm 2}Kling Team, Kuaishou Technology\\
}

\usepackage{bibentry}

\begin{document}

\maketitle


\begin{abstract}
Current video generation models perform well at single-shot synthesis but struggle with multi-shot videos, facing critical challenges in maintaining character and background consistency across shots and flexibly generating videos of arbitrary length and shot count. To address these limitations, we introduce \textbf{FilmWeaver}, a novel framework designed to generate consistent, multi-shot videos of arbitrary length. First, it employs an autoregressive diffusion paradigm to achieve arbitrary-length video generation. To address the challenge of consistency, our key insight is to decouple the problem into inter-shot consistency and intra-shot coherence. We achieve this through a dual-level cache mechanism: a shot memory caches keyframes from preceding shots to maintain character and scene identity, while a temporal memory retains a history of frames from the current shot to ensure smooth, continuous motion. The proposed framework allows for flexible, multi-round user interaction to create multi-shot videos. Furthermore, due to this decoupled design, our method demonstrates high versatility by supporting downstream tasks such as multi-concept injection and video extension. To facilitate the training of our consistency-aware method, we also developed a comprehensive pipeline to construct a high-quality multi-shot video dataset. Extensive experimental results demonstrate that our method surpasses existing approaches on metrics for both consistency and aesthetic quality, opening up new possibilities for creating more consistent, controllable, and narrative-driven video content. 
Project Page: \url{https://filmweaver.github.io}
\end{abstract}
\section{Introduction}
With the continuous advancement of video diffusion models~\cite{svd,hunyuanvideo,wan}, video generation systems have demonstrated remarkable capabilities and found applications across various domains~\cite{xue2024human}. However, these models are primarily designed for single video generation. Compared to conventional single-shot video generation, multi-shot videos offer significantly higher practical value in filmmaking, storytelling, and other creative fields, as they enable the construction of more complex narratives. Nevertheless, this task presents greater challenges, primarily in two aspects: i) maintaining consistency of the same subjects or backgrounds across different shots, and ii) managing shot duration and the number of shots. These requirements cannot be directly addressed by traditional video generation models.

The most straightforward approach to multi-shot generation involves using multiple prompts to generate each shot independently and then concatenating them. However, the coarse-grained textual representations struggle to ensure consistency between shots. To address the consistency problem in multi-shot generation, early methods typically employed complex pipelines involving multiple models~\cite{videodirectorgpt, videostudio,mora, xiao2025captain}, decomposing multi-shot video generation into keyframe generation and image-to-video synthesis. These approaches achieved multi-shot consistency through object insertion techniques or attention reference mechanisms~\cite{storydiffusion} during keyframe generation. However, such methods rely heavily on complex pipeline designs and, due to the independent generation of individual segments without considering global temporal coherence, often result in visual discontinuities and abrupt scene transitions.

Recent work has proposed simultaneous multi-shot generation methods that partition the video generated by a video generation model into multiple segments~\cite{shotadapter,mask2dit,lct}, with each segment corresponding to a shot. These methods are trained on corresponding multi-shot datasets and achieve improved inter-shot consistency. However, since multiple shots share a single sequence, the duration of individual shots is severely constrained. TTT~\cite{TTT} introduces RNN-like mechanisms in the intermediate layers of DiT~\cite{dit} to maintain shot consistency, but these lack long-term memory, have fixed shot durations, and incur substantial training costs. LCT~\cite{lct} employs additional positional encoding to achieve multi-shot generation and shot extension, but it requires two-stage training and only supports the MM-DiT~\cite{cogvideox} architecture.


%

To address these limitations, we introduce a novel cache-guided autoregressive framework for multi-shot video generation. Our approach is founded on a key insight: the explicit decoupling of inter-shot consistency and intra-shot coherence. To this end, we design a dual-level cache system composed of a Shot Cache and a Temporal Cache. The Shot Cache is responsible for maintaining inter-shot consistency; it stores identifying visual features of subjects and backgrounds from preceding shots. When generating a new shot, it allows the model to retrieve relevant context based on the prompt, ensuring character and style continuity across narrative breaks. In parallel, the Temporal Cache manages intra-shot coherence by retaining a memory of the immediately preceding frames within the current shot, guaranteeing fluid motion and preventing visual flickering.

\begin{figure}[t]
    \centering
    \includegraphics[width=\linewidth]{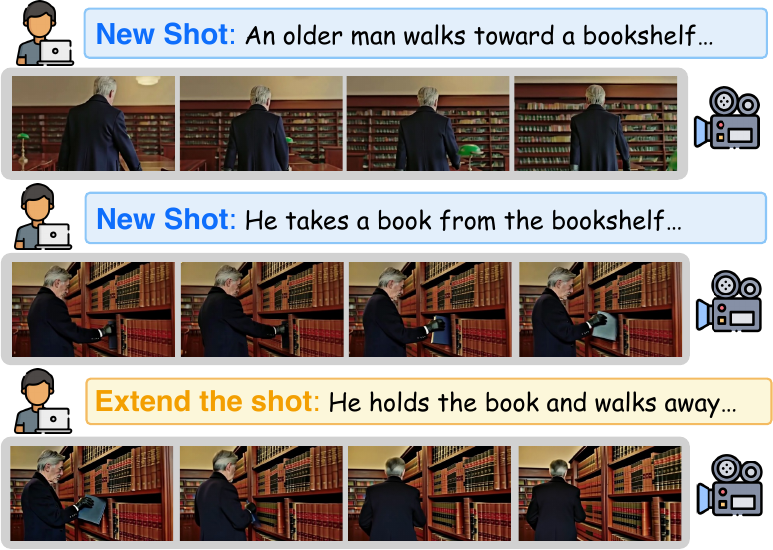}
    \caption{Our framework supports interactive creation of multi-shot sequences. Users can generate new shots or extend existing ones to build a coherent narrative.}
    \label{fig:intro}
\end{figure}
Our framework integrates this dual-level cache system using in-context injection, a mechanism that avoids architectural modifications and thus ensures broad compatibility with existing pre-trained text-to-video models. As shown in Figure~\ref{fig:intro}, this approach provides intuitive interactive control over the narrative flow. 
This core flexibility is the key that unlocks not only the generation of videos with an arbitrary number of shots and durations but also a range of challenging downstream tasks, such as multi-concept injection~\cite{huang2025conceptmaster, canonswap} and interactive video extension.

Training this model requires constructing high-quality consistent multi-shot video datasets, which are currently scarce. We propose a data collection pipeline for building high-quality multi-shot datasets, incorporating shot-to-scene shot collection and scene-to-shot multi-level annotation refinement to ensure high-quality consistent caption construction. Given the lack of multi-shot video evaluation methods, we propose evaluation metrics and several baselines based on identity consistency, background consistency, and video quality. In summary, our contributions are as follows:

\begin{itemize}
    \item We introduce a novel cache-guided autoregressive framework for multi-shot video generation. This architecture uniquely integrates a dual-level cache into the autoregressive flow, enabling the synthesis of coherent videos of arbitrary length by explicitly managing intra and inter shot consistency.
    \item The proposed framework exhibits remarkable flexibility, supporting a range of challenging downstream tasks including multi-concept character injection, interactive video extension, showcasing its broad applicability.
    \item We propose a high-quality consistent shot collection pipeline that includes shot segmentation and clustering, and through multi-level agent annotation, we collect a high-quality multi-shot dataset.
    \item Extensive experiments demonstrate that our framework surpasses existing methods in video quality and character/background consistency.
\end{itemize}

\section{Related Works}
\subsection{Single-shot Video Generation}
\subsubsection{Video Diffusion}
Video diffusion models~\cite{svd,hunyuanvideo,wan,cogvideox} have evolved from their successful applications in image synthesis, with the core challenge being modeling temporal consistency. Early methods~\cite{svd,animatediff} leverage image generation model priors by inserting additional temporal attention modules to learn motion dynamics, but the video quality is limited. Some approaches~\cite{sv3d, grid} partition the image into a grid of patches to achieve consistent multi-view or video generation. Current video generation models~\cite{hunyuanvideo,wan} typically employ a 3D-DiT architecture that flatten video height, width, and temporal dimensions into a single dimension for attention computation, further improving generated video quality. 

\subsubsection{Long Video Generation}
Extending video generation beyond short clips remains a significant challenge. Early works employ various strategies, such as leveraging latent diffusion models~\cite{ldm}, generating variable-length videos from text sequences, and implementing coarse-to-fine architectures~\cite{pyramidal}. Other efforts focus on achieving temporal consistency through distributed generation or extending pre-trained models via noise rescheduling techniques~\cite{ditctrl}.
Recently, research focus has shifted toward long-context video modeling to better utilize historical information~\cite{diffusionforcing,skyreels}. FAR~\cite{far} proposed a framework conditioned on both long-term and short-term context windows, while FramePack~\cite{framepack} introduces hierarchical compression of context frames. These methods highlight the critical role of context as a form of memory in achieving scene-consistent long video generation. Unlike these approaches, in multi-shot scenarios, we propose a two-level cache system to simultaneously achieve inter-shot consistency and intra-shot coherence.
\subsection{Multi-shot Generation}
Multi-shot generation, particularly in narrative contexts, faces challenges in maintaining consistency. Early works in story visualization center on image generation. To this end, various techniques are developed; for instance, AutoStudio~\cite{autostudio}  employ IP-Adapter~\cite{ipadapter} for character identity injection, while StoryDiffusion~\cite{storydiffusion} and Consistory~\cite{consistory} utilizes attention concatenation mechanisms~\cite{luo2025object} for visual consistency.
Early multi-shot video generation methods adopt a two-stage paradigm of keyframe generation followed by I2V video generation. Mora~\cite{mora} and MovieAgent~\cite{movieagent} adopt multi-agent frameworks, while VideodirectorGPT~\cite{videodirectorgpt} introduces layout-to-video generation and VideoStudio~\cite{videostudio} distinguish foreground and background with action information. However, these complex, multi-model paradigms generate segments independently, which often leads to visual discontinuities.

Recent methods enhance inter-shot consistency through simultaneous multi-shot generation by decomposing the generated sequence into multiple shots~\cite{mask2dit,shotadapter}. However, splitting the original video length results in very short duration per shot, reducing practical applicability. TTT~\cite{TTT} introduces RNNs at intermediate DiT layers but lacks long-range dependencies and incurs high training costs. LCT~\cite{lct} and EchoShot~\cite{echoshot} employ complex positional encoding to distinguish shots. In contrast, we adopt simple positional encoding with a two-level cache system to simultaneously ensure inter-shot consistency and intra-shot coherence through autoregressive single-step training.

\section{Method}

Our proposed framework, FilmWeaver, introduces an autoregressive video generation paradigm capable of producing multi-shot videos of arbitrary length and shot count. The core of our method is a novel dual-level cache mechanism designed to maintain both intra-shot coherence and inter-shot consistency. This is complemented by a strategic two-stage training process and a meticulous data curation pipeline to enable robust learning.
\begin{figure}
    \centering
    \includegraphics[width=\linewidth]{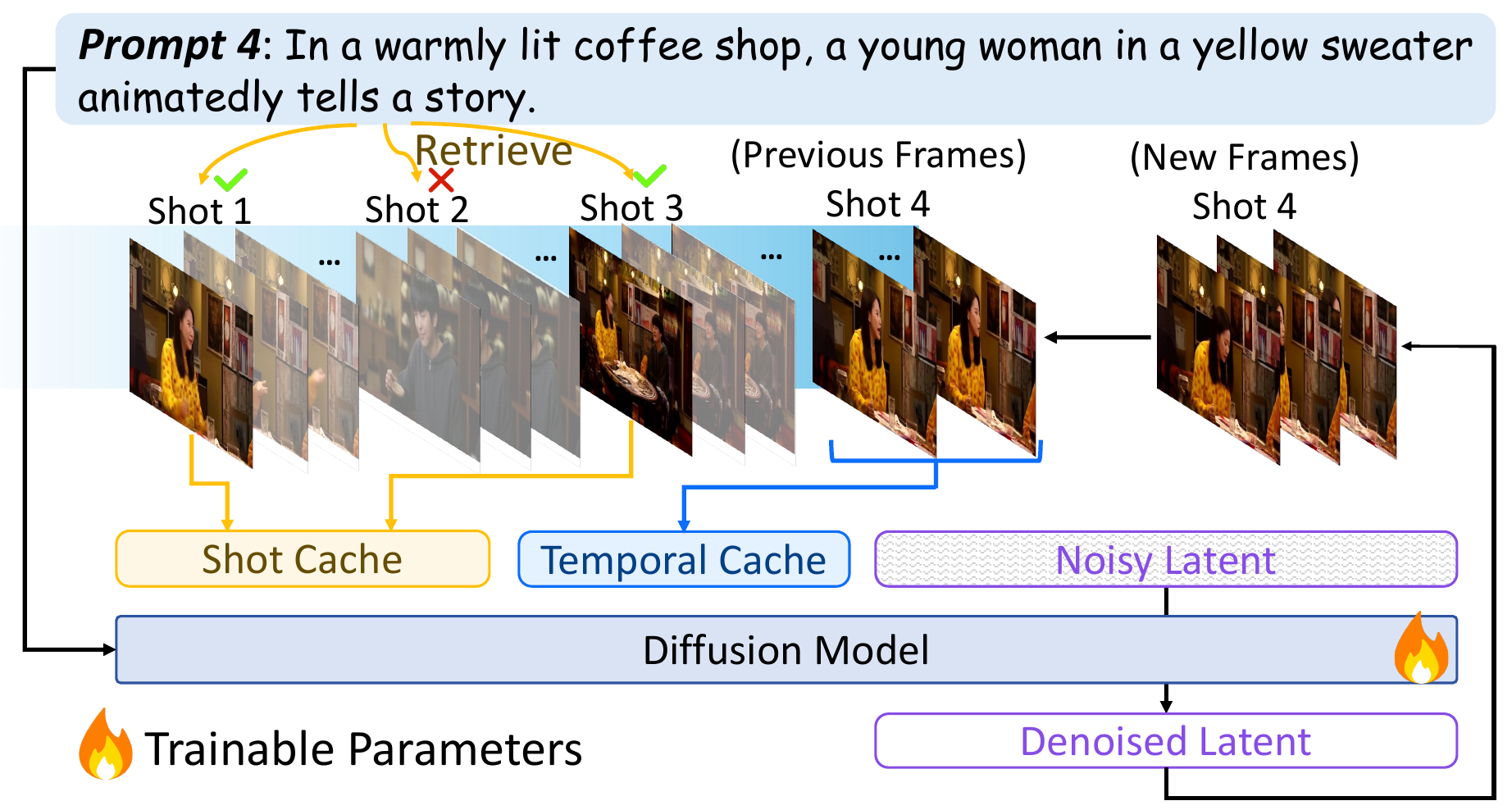}
    \caption{The framework of FilmWeaver. New video frames are generated autoregressively and consistency is enforced via a dual-level cache mechanism: a Shot Cache for long-term concept memory, populated through prompt-based key-frames retrieval from past shots, and a Temporal Cache for intra-shot coherence.}
    \label{fig:pipeline}
\end{figure}

\subsection{Autoregressive Generation with Dual-Level Cache}
As illustrated in Figure~\ref{fig:pipeline}, our framework's central innovation is a dual-level cache that provides contextual information to the diffusion model. This framework comprises two key components: a Temporal Cache, which serves as a short-term memory to ensure the fluid coherence of generated content within a single shot, and a Shot Cache, which provides long-term memory to maintain the consistency of core concepts, such as characters and backgrounds across different shots.
Both caches are injected into the model via in-context learning. This modifies the conditioning of the denoiser, which now takes the text prompt $\mathbf{c}_{\text{text}}$, the Temporal Cache $C_{\text{temp}}$, and the Shot Cache $C_{\text{shot}}$ as input. The model is thus trained with the following objective:
\begin{equation}
    \mathcal{L} = \mathbb{E}_{\mathbf{v}_0, \mathbf{c}_{\text{text}}, \epsilon, t} \left[ \left\| \epsilon - \epsilon_\theta(\mathbf{v}_t, t, \mathbf{c}_{\text{text}}, C_{\text{temp}}, C_{\text{shot}}) \right\|^2 \right].
\end{equation}
This approach conditions the generation process on relevant past information without altering the model architecture.

\paragraph{Temporal Cache for Intra-Shot Coherence.}
To ensure seamless continuity within a single shot, we employ a Temporal Cache. This cache functions as a sliding window, storing conditioning frames from the immediate past of the current generation chunk. As new frames are generated, they are added to the cache, while the oldest frames are discarded. Given the high temporal redundancy in videos, storing all past frames is computationally prohibitive. Therefore, inspired by recent works~\cite{far, framepack}, we implement a differential compression strategy. Frames closer to the current generation window are preserved with higher fidelity, while frames further in the past are progressively compressed. This approach efficiently retains essential motion and content information while minimizing computational overhead.


\paragraph{Shot Cache for Inter-Shot Consistency.}
To ensure inter-shot consistency, we introduce the Shot Cache. When generating a new shot, this cache is built by selecting the top-K keyframes from previous shots that are most semantically relevant to the new text prompt. Relevance is determined by computing the cosine similarity between the CLIP embeddings~\cite{clip} of the prompt and each candidate keyframe. The resulting cache provides a concise yet highly relevant visual summary of the narrative history, guiding the model to maintain consistency. The retrieval process is formulated as:
\begin{equation}
    C_{\text{shot}} = \underset{kf \in \mathcal{KF}}{\operatorname{arg\,top-k}} \left( \operatorname{sim}(\phi_T(\mathbf{c}_{\text{text}}), \phi_I(kf)) \right),
\end{equation}
where $C_{\text{shot}}$ is the resulting Shot Cache, $\mathcal{KF}$ is the set of all keyframes ($kf$) from previous shots, and $\mathbf{c}_{\text{text}}$ is the text prompt for the new shot. Furthermore, $\phi_T$ and $\phi_I$ are the pretrained CLIP text and image encoders, respectively, $\operatorname{sim}(\cdot, \cdot)$ denotes the cosine similarity function, and the $\operatorname{arg\,top-k}$ operator selects the K keyframes with the highest similarity scores. 


\begin{figure*}[ht]
    \centering
    \includegraphics[width=0.95\linewidth]{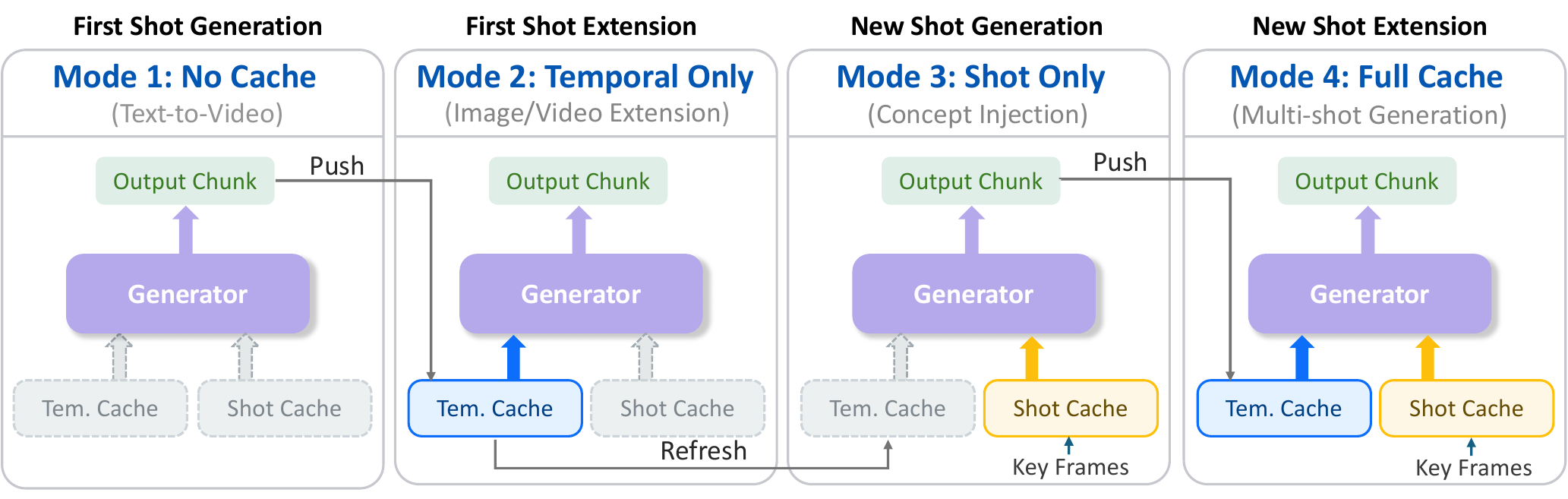}
    \caption{The multi-shot generation process involves four distinct scenarios, each corresponding to a different inference mode: (1) No Cache initializes generation and populates the caches; (2) Temporal Only extends the current shot using temporal context; (3) Shot Only transitions to a new shot by injecting key frames from previous shots; and (4) Full Cache combines both contexts.}
    \label{fig:inference}
\end{figure*}

\subsubsection{Inference Stages and Modes}
\label{sec: modes}
As illustrated in Figure~\ref{fig:inference}, our multi-shot generation process unfolds across four sequential stages. Each stage is governed by a distinct inference mode, determined by the state of our dual-level cache, enabling versatile and consistent video creation.

\begin{itemize}
    \item \textbf{First Shot Generation (No Cache).} The process begins with both caches empty ($C_{\text{temp}} = \emptyset, C_{\text{shot}} = \emptyset$). The model operates as a standard text-to-video (T2V) generator, creating the initial video chunk from the text prompt and populating the caches for subsequent stages.

    \item \textbf{First Shot Extension (Temporal Only).} To generate subsequent chunks within the same shot, the model leverages the populated Temporal Cache ($C_{\text{temp}} \neq \emptyset, C_{\text{shot}} = \emptyset$). This ensures high temporal coherence and fluid motion, ideal for applications like video extension or image-to-video (I2V) generation.

    \item \textbf{New Shot Generation (Shot Only).} To transition to a new shot, the Temporal Cache is cleared while the Shot Cache is populated with keyframes from previous shots ($C_{\text{temp}} = \emptyset, C_{\text{shot}} \neq \emptyset$). This mode generates a new scene that maintains long-term consistency of key elements like characters and backgrounds. We can also artificially set reference frames to achieve multiple concept injection.

    \item \textbf{New Shot Extension (Full Cache).} When extending the new shot, both caches are active ($C_{\text{temp}} \neq \emptyset, C_{\text{shot}} \neq \emptyset$). The generator leverages the Temporal Cache for immediate coherence and the Shot Cache for long-term consistency, seamlessly blending both contexts.
\end{itemize}

Our training process is designed to accommodate all four scenarios, ensuring the model operates robustly across the entire multi-shot generation sequence.

\subsection{Training Strategy}
To simplify the learning task and promote stable convergence, we adopt a two-stage training strategy, coupled with data augmentation techniques to reduce the model's over-reliance on the cache.

\paragraph{Progressive Training Curriculum.}
Our strategy unfolds in two progressive stages. The first stage focuses exclusively on teaching the model to generate long, coherent single-shot videos with only text input. During this phase, the Shot Cache is disabled (i.e., its inputs are zeroed out), and the model is trained only with the Temporal Cache. This foundational step allows the model to master intra-shot dynamics without the added complexity of cross-shot consistency. Building on this foundation, the second stage activates the Shot Cache and fine-tunes the model on a mixed curriculum that includes all four cache-based scenarios. This progressive approach, where the model is pre-trained on the simpler long-video task, allows it to tackle the more complex challenge of multi-shot consistency with greater efficiency and faster convergence.

\paragraph{Data Augmentation.}
We observe that during training, the model develops an over-reliance on the provided visual context, leading to a ``copy-paste" phenomenon. This behavior  significantly reduces motion dynamism and diminishes the model's adherence to the textual prompt. To mitigate this, we employ several targeted regularization techniques. First, we apply negative sampling to the Shot Cache by randomly introducing irrelevant keyframes during training. This compels the model to discriminate between useful and distracting context, guided by the prompt. Furthermore, inspired by prior strategies, we introduce noise to both caches to discourage exact replication. However, we find that excessive noise in the Temporal Cache degrades video coherence. We therefore adopt an asymmetric noising strategy: a substantial noise level, corresponding to a random diffusion timesteps between 100 to 400, is applied to the Shot Cache, while a much milder noise level (0--100 timesteps) is applied to the Temporal Cache. This dual approach effectively curbs over-reliance on visual context, thereby enhancing the ability of prompt following and generative quality.

\begin{figure*}
    \centering
    \includegraphics[width=\linewidth]{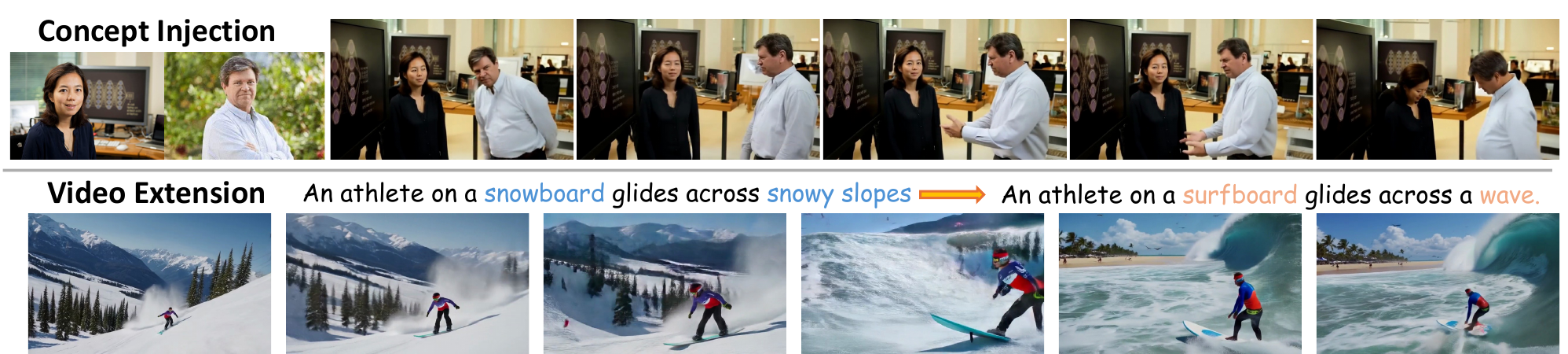}
    \caption{Examples of FilmWeaver's downstream applications.  Our framework enables multi-concept Injection (Top), generating dynamic videos from reference images while faithfully preserving subject identity. It also excels at complex video extension tasks (Bottom), where the narrative is steered mid-sequence by modifying a text prompt. The seamless transition from a ``snowboard" to a ``surfboard" scene, while maintaining the athlete's appearance.}
    \label{fig:application}
\end{figure*}
\begin{figure}[h]
    \centering
    \includegraphics[width=\linewidth]{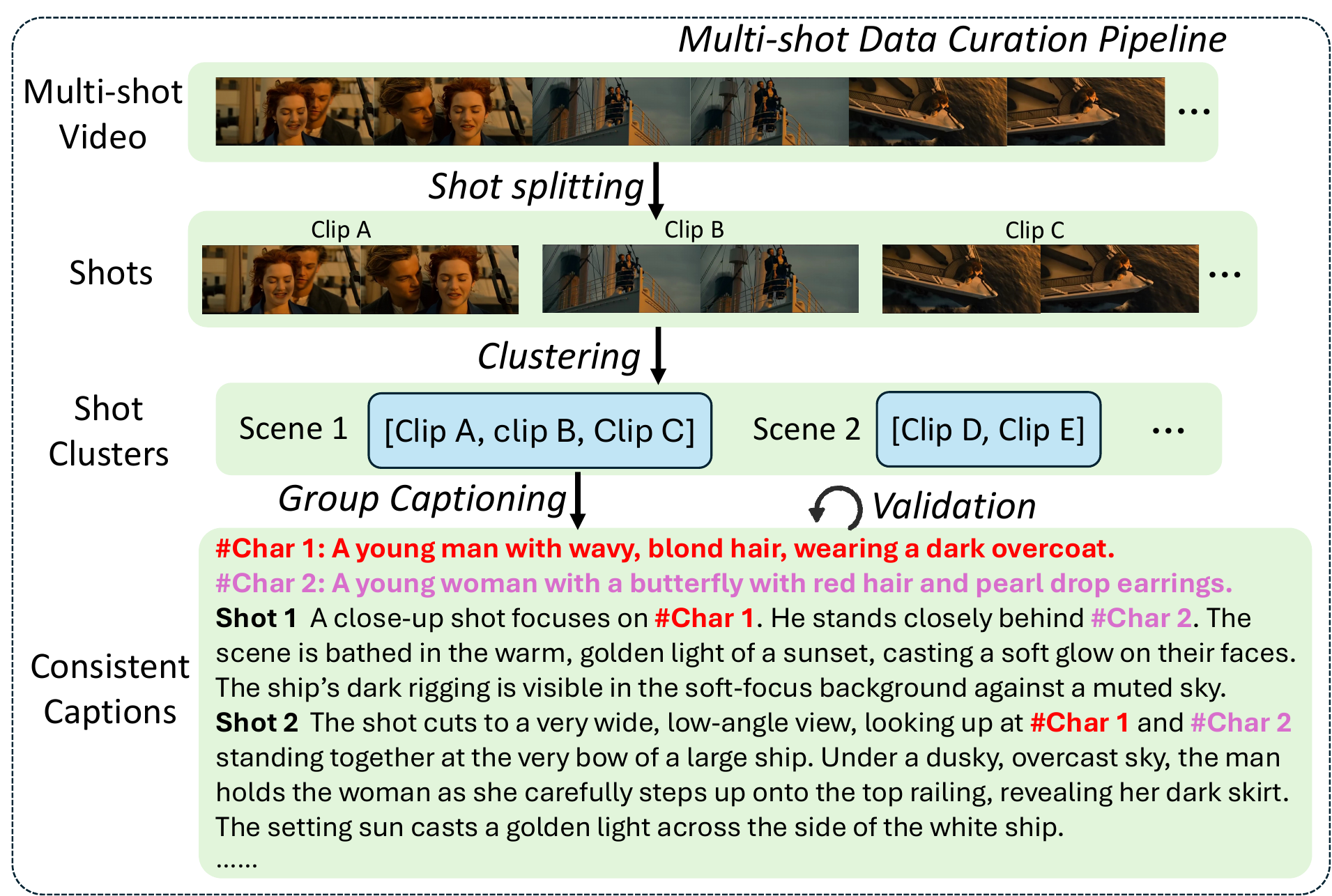}
    \caption{The pipeline of Multi-shot data curation, which first segments videos into shots and clusters them into coherent scenes. We then introduce a Group Captioning strategy that jointly describes all shots within a scene, enforcing consistent attributes for characters and objects. This process, finalized with a validation step, yields a high-quality dataset of video-text pairs with strong temporal coherence.}
    \label{fig:curation}
\end{figure}

\subsection{Multi-Shot Data Curation}
A major obstacle in multi-shot video generation is the lack of high-quality, consistently annotated datasets. To address this, we developed a comprehensive data curation pipeline as illustrated in Figure~\ref{fig:curation}. For the initial Shot Splitting stage, we employ an expert model~\cite{panda} to segment the source videos into individual shots. 
These shots are then grouped into coherent multi-shot sequences through Scene Clustering, a process that utilizes a sliding window to evaluate the CLIP similarity between clips.
Following clustering, scenes undergo a rigorous filtering protocol to ensure data quality, removing clips shorter than one second and scenes featuring more than three distinct individuals.
Our annotation pipeline begins with a Group Captioning strategy. We provide the entire sequence of shots from a scene to Gemini 2.5 Pro, which generates descriptions for all shots simultaneously. This joint-captioning approach is can maintain consistent descriptions of the same character across different shots, while also accurately reflecting cinematic changes like transitions from a wide shot to a close-up. Subsequently, each generated annotation undergoes a Validation step. In this stage, each individual shot and its corresponding caption are fed back into the model to verify its accuracy and refine any ambiguous or generic phrasing. 

\section{Extended Applications}

The architecture of FilmWeaver enables a versatile range of downstream applications. As illustrated in Figure~\ref{fig:application}, our framework's can easily handle multi-concept injection and video extension. 
First, for multi-concept injection (Top), we manually populate the Shot Cache with provided concept images, which denotes the Mode 3 in Figure~\ref{fig:inference}. FilmWeaver can then generate a coherent, dynamic scene that seamlessly integrates these multiple subjects while faithfully preserving their individual identities.
Furthermore, the framework excels at Dynamic Video Extension (Bottom). This is achieved by modifying the text prompt mid-sequence while preserving the Temporal Cache to maintain continuity. This process corresponds to Mode 2 (Temporal Only) in Figure~\ref{fig:inference}, allowing the narrative to be steered on the fly. For instance, it can fluidly transition a video from the context of ``an athlete snowboarding on snowy slopes'' to ``surfing on ocean waves'' in direct response to the new prompt. The model's ability to leverage its distinct cache modes for such sophisticated tasks highlights its potential for advanced video editing and creative narrative exploration.

\section{Experiments}

\subsection{Implementation Details}
For our implementation, we adapt the HunyuanVideo~\cite{hunyuanvideo}. The batch size is set to 16. The first stage is trained for 10K steps, and the second stage is trained for 10K steps.

\begin{figure*}[h]
    \centering
    \includegraphics[width=0.94\linewidth]{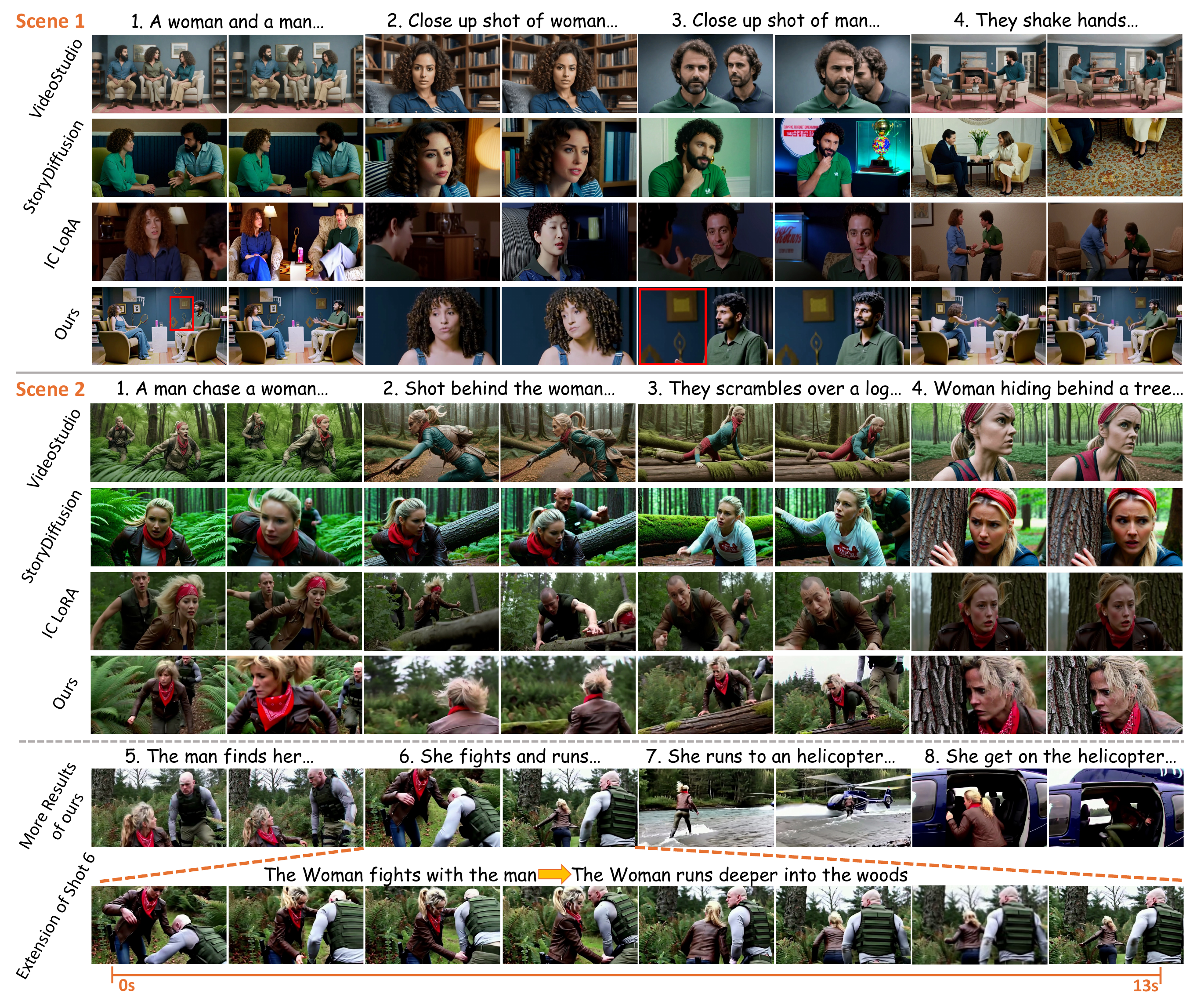}
    \caption{ Qualitative comparison of multi-shot video generation. We compare two scenarios where existing methods tend to confuse character features and fail to maintain background consistency, while our method handles these challenges effectively. Additionally, we showcase more shots generated by our framework in Scene 2, forming a continuous narrative. For the sixth shot in Scene 2, we demonstrate our video extension capability by creating a coherent long video using two prompts.}
    \label{fig:qualitative}
\end{figure*}

\subsection{Comparison with Existing Methods}
To evaluate FilmWeaver's performance, we compare it against two main categories of existing multi-shot video generation methods. The first category utilized the full pipeline to generate multi-shot videos, as represented by VideoStudio~\cite{videostudio}. The second category denoted key-frames based methods comprises two-stage pipelines that first generate keyframes and then animate them using an I2V model. For these pipelines, we use StoryDiffusion~\cite{storydiffusion} and IC-LoRA~\cite{iclora} for consistent keyframe generation, followed by the powerful Hunyuan I2V model for the animation step. Since IC-LoRA generates a limited number of images per run, we invoke it multiple times to produce the required keyframes. Given that VideoStudio and IC-LoRA require structured inputs, we use Gemini 2.5 Pro~\cite{gemini} to reformat the prompts accordingly.

\paragraph{Qualitative Comparison.}
We present a comprehensive qualitative comparison in Figure~\ref{fig:qualitative}, evaluating our method against leading approaches across two distinct and challenging narrative scenarios.
In Scene 1, a conversation scene with alternating wide and close-up shots, competing methods exhibit severe consistency failures. They struggle with multi-person identity preservation, leading to feature entanglement where attributes like clothing and facial structures are blended between characters and their backgrounds are inconsistent. In contrast, our method successfully preserves the distinct appearance of each individual and maintains a stable background across all shots. As highlighted by the red box in the fourth line, details such as the wall art behind the man in Shot 3 remain perfectly consistent with the Shot 1.
Similarly, our method robustly preserves character identities during the dynamic action in Scene 2, overcoming the appearance inconsistencies common to competitor models.

Crucially, our framework's capabilities extend far beyond short, four-shot sequences. We demonstrate this by generating a continuous and coherent 8-shot narrative for Scene 2, showcasing its strength in long-form storytelling. Moreover, we explicitly highlight our video extension capability within ``Extension of Shot 6," where we use a second prompt to seamlessly continue the action within a single shot, creating a longer and more dynamic clip. These results underscore our model's superior ability to generate consistent, controllable, and lengthy narrative-driven video content. For additional examples, please refer to the appendix.

\begin{table}[htbp]
\centering
\small 
\setlength{\tabcolsep}{2pt} 
\begin{tabular}{lcccccc}
\toprule
\multirow{2}{*}{Method} & \multicolumn{2}{c}{Visual Quality} & \multicolumn{2}{c}{Consistency(\%)} & \multicolumn{2}{c}{Text Alignment} \\
\cmidrule(lr){2-3} \cmidrule(lr){4-5} \cmidrule(lr){6-7}
 & Aes.$\uparrow$ & Incep.$\uparrow$ & Char.$\uparrow$ & All.$\uparrow$ & Char. $\uparrow$ & All. $\uparrow$ \\
\midrule
VideoStudio & 32.02 & 6.81 & \underline{73.34} & 62.40 & 20.88 & \textbf{31.52} \\
StoryDiffusion & \textbf{35.61} & \underline{8.30} & 70.03 & 67.15 & 20.21 & 30.86 \\
IC-LoRA & 31.78 & 6.95 & 72.47 & \underline{71.19} & \underline{22.16} & 28.74\\
Ours & \underline{33.69} & \textbf{8.57} & \textbf{74.61} & \textbf{75.12} & \textbf{23.07} & \underline{31.23}\\
\bottomrule
\end{tabular}
\caption{Quantitative comparison with existing methods.}
\label{tab:quantitative}
\end{table}

\paragraph{Quantitative Comparison.}
These visual observations are substantiated by our quantitative evaluation, summarized in Table~\ref{tab:quantitative}. For quantitative assessment, we evaluate performance across three key dimensions: Visual Quality, Consistency, and Text Alignment. For Visual Quality, we adopt the evaluation protocol from MovieBench~\cite{moviebench}, which contains Aesthetics Score (Aes.) and Inception Score (Incep.). For Consistency and Text Alignment, we report metrics for both character-specific (Char.) and overall (All.) aspects.

Quantitative results in Table 1 validate our method's superiority. We achieve state-of-the-art results in both Consistency metrics and character-level Text Alignment. Crucially, our method also attains the highest Inception Score, demonstrating top-tier visual quality, and remains highly competitive across all other metrics.
We posit that the visual quality of our method can be further enhanced through improved data curation and optimized training strategies.

\begin{table}[htbp]
\centering
\small 
\setlength{\tabcolsep}{4pt} 
\begin{tabular}{lcccccc}
\toprule
\multirow{2}{*}{Method} & \multicolumn{2}{c}{Visual Quality} & \multicolumn{2}{c}{Consistency(\%)} & \multicolumn{2}{c}{Text Alignment} \\
\cmidrule(lr){2-3} \cmidrule(lr){4-5} \cmidrule(lr){6-7}
 & Aes.$\uparrow$ & Incep.$\uparrow$ & Char.$\uparrow$ & All.$\uparrow$ & Char. $\uparrow$ & All. $\uparrow$ \\
\midrule
w/o A & 30.04 & 7.77 & 72.36 & 75.92 & 21.88 & 28.12 \\
w/o S & 33.92 & 8.63 & 68.11 & 65.44 & 22.41 & 31.79 \\
w/o T & 31.61 & 8.36 & 70.79 & 70.57 & 20.21 & 30.70 \\
Ours & 33.69 & 8.57 & 74.61 & 75.12 & 23.07 & 31.23\\
\bottomrule
\end{tabular}
\caption{Quantitaive results of ablation study.}
\label{tab:ablation}
\end{table}

\subsection{Ablation Studies}
We conduct ablation studies to validate our core components: the Shot Cache (S), the Temporal Cache (T), and the {noise augmentation (A) strategy. 
\begin{figure}[h]
    \centering
    \includegraphics[width=\linewidth]{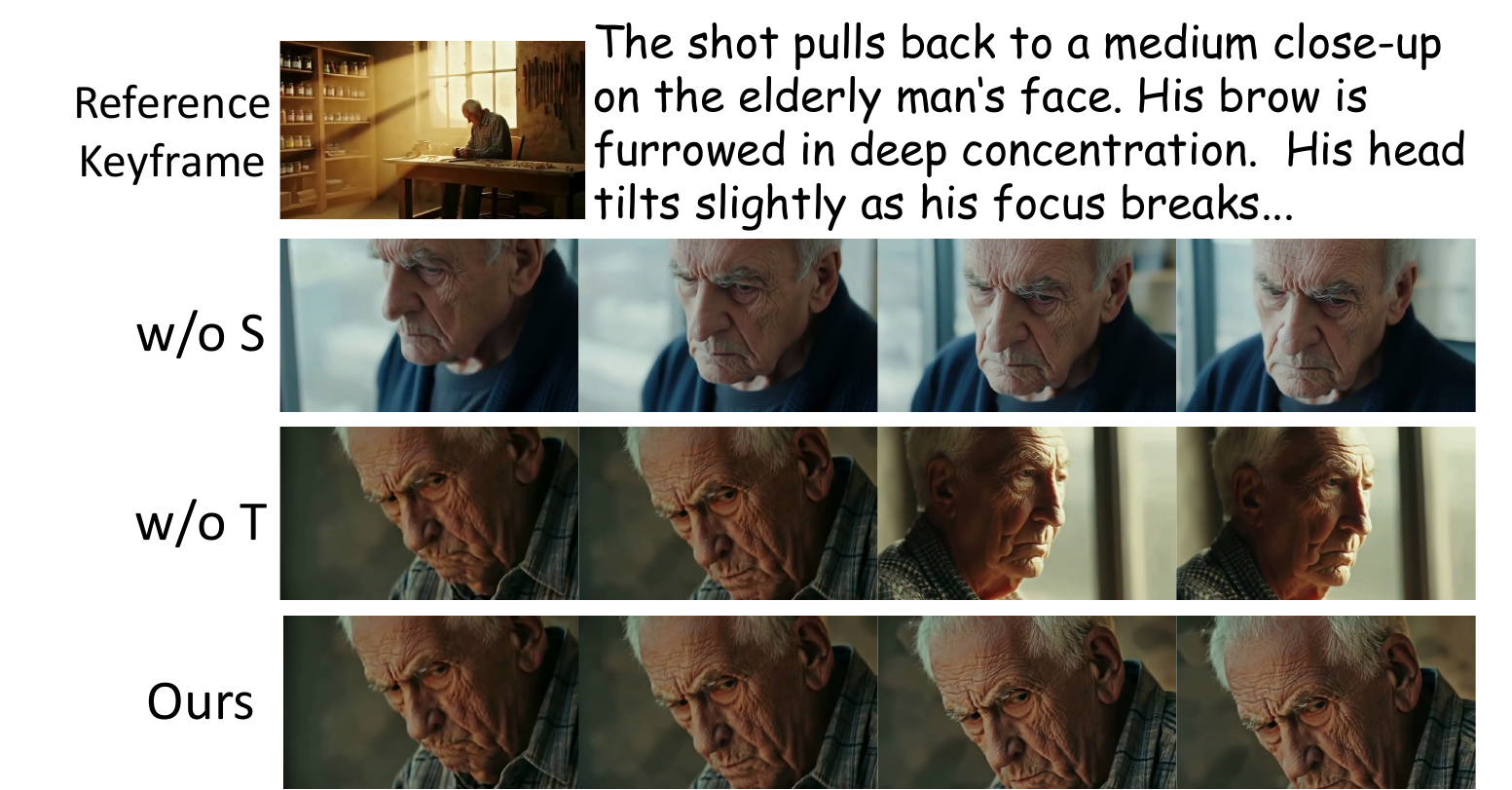}
    \caption{Qualitative ablation study of our dual-level cache. Without the shot cache (w/o S), the model fails to maintain visual style and the clothes of character. Without the temporal cache (w/o T), the generated sequence lacks coherence, resulting in disjointed motion. Our full method successfully preserves both appearance and motion continuity.}
    \label{fig:ablation2}
\end{figure}
As shown in Figure~\ref{fig:ablation2}, removing the Shot Cache (w/o S) leads to a severe loss of long-term visual consistency, failing to preserve character appearance and scene style. Removing the Temporal Cache (w/o T) results in temporally incoherent and disjointed motion. Furthermore, Figure~\ref{fig:ablation} demonstrates that without noise augmentation (w/o A), the model's over-reliance on visual context from the Temporal Cache hinders its ability to adapt to dynamic prompts, thus reducing prompt controllability during video extension.
Table~\ref{tab:ablation} quantitatively corroborates these findings. The variants without the caches show a significant drop in consistency metrics, while the absence of noise augmentation leads to a notable decrease in text alignment scores. These results confirm that each component is essential: the Shot Cache for inter-shot consistency, the Temporal Cache for intra-shot coherence, and noise augmentation for robust prompt adherence.

\begin{figure}
    \centering
    \includegraphics[width=\linewidth]{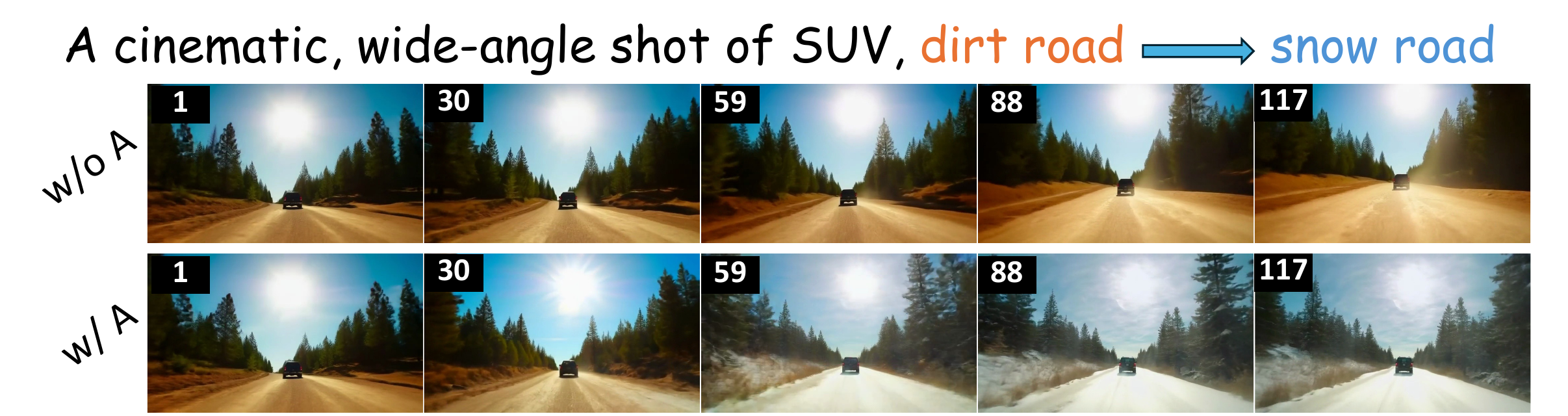}
    \caption{Qualitative ablation study on noise augmentation. Without noise augmentation, the model over-relies on past frames, hindering the ability of prompt following, which is crucial in video extension. Applying noise reduces this dependency and improves the ability of prompt following.}
    \label{fig:ablation}
\end{figure}

\section{Conclusion}
We present FilmWeaver, a novel cache-guided autoregressive framework addressing consistency and duration in multi-shot video generation. Our key contribution is a dual-level cache mechanism combining a retrieval-based Shot Cache for long-term consistency with a Temporal Cache for intra-shot motion coherence. Supported by noise augmentation for better prompt adherence, this architecture enables generating coherent videos of arbitrary length. Experiments on a curated dataset show FilmWeaver significantly outperforms existing methods in visual consistency and aesthetic quality. Its flexibility allows direct application to tasks like video extension. FilmWeaver represents a substantial advance in creating complex, controllable narrative videos, opening new possibilities for automated storytelling.

\bibliography{aaai2026}

\clearpage
\appendix
\section{Dual-Level Cache}
In this section, we show more details of our proposed dual-level cache.
\subsection{Details of Dual-Level Cache}
In our method, we employ a dual-level cache system comprising a Shot Cache and a Temporal Cache. The Shot Cache maintains inter-shot consistency by storing keyframes from previous shots, while the Temporal Cache ensures intra-shot coherence using a sliding window of the most recent frames from the current, unfolding shot. In practice, both caches store not the raw pixels, but the latent representations encoded by a VAE.

For the Shot Cache, we set K to 3 based on a crucial trade-off between performance and efficiency. While using more keyframes  can provide additional visual context, we found the improvement in generation quality becomes minimal. Conversely, the costs in terms of computation and training complexity continue to rise. As demonstrated in our qualitative analysis (see Figure~\ref{fig:retrieval}), a cache of this size is sufficient to capture the diverse concepts required to maintain consistency in complex, multi-shot generation scenarios.
For the Temporal Cache, our approach is guided by the intuition that historical frames closer to the current frame are more important than those further away. Therefore, following recent autoregressive methods, we adopt a hierarchical compression strategy. We divide the sliding window's compression into three tiers: the most recent latent is kept uncompressed, the next two latents undergo 4x compression, and the final 16 latents are compressed by a factor of 32. This method effectively manages token consumption as the number of latents increases. This compressed sliding window design significantly reduces computational overhead.
In our setup, a single autoregressive step generates 6 latents, which is equivalent to 24 video frames. At a standard 24 FPS frame rate, this configuration allows the length of each shot to be controlled in precise, one-second increments.
\subsection{Additional Ablation Results}
To further validate the effectiveness of our Shot Cache, we present a case study that isolates its core function of concept injection.
In Figure~\ref{fig:supp_ablation}, both generated videos use the Shot Cache, but are given with different references. When the reference includes a frontal view of an older man (A), the cache successfully injects this identity into the video, maintaining it consistently. When the reference lacks this concept (B), the generated video create an identity.
This direct correspondence between the reference concept and the generated output serves as clear proof of the Shot Cache's role. It confirms the cache operates as designed, acting as the primary mechanism for enforcing identity consistency from a given source.
\begin{figure}
    \centering
    \includegraphics[width=\linewidth]{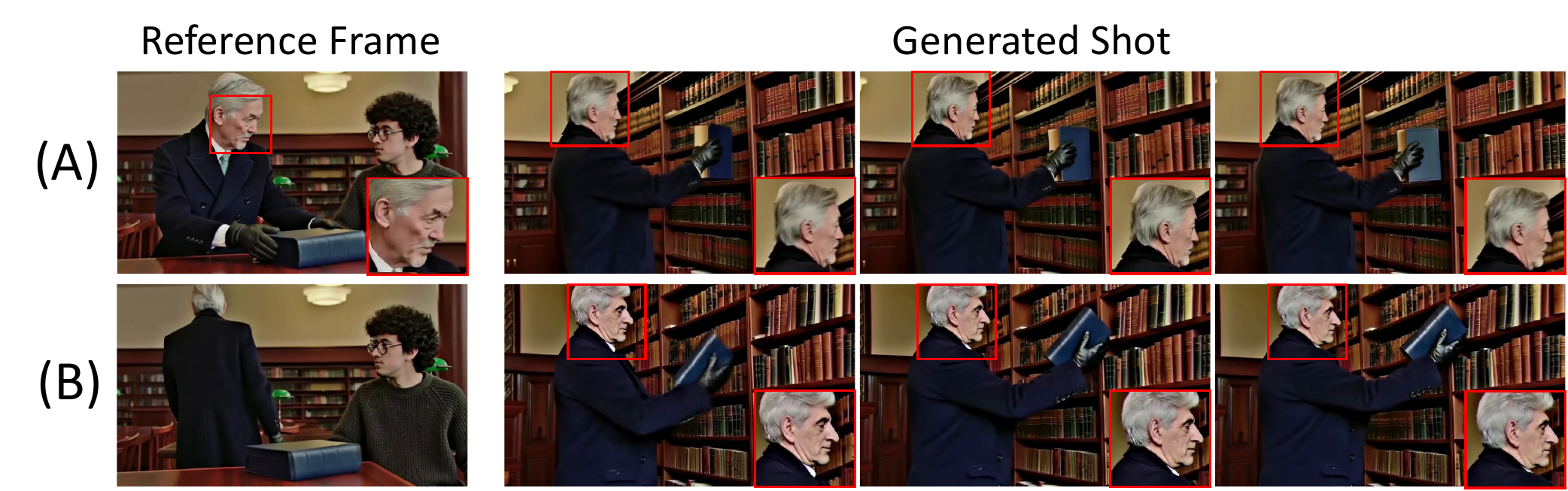}
    \caption{Effectiveness of the Shot Cache demonstrated via an additional ablation. In (A), the generated shot correctly inherits the character identity from the reference image. In (B), without the character ID in the reference, a new identity is created. This further validates the effectiveness of our Shot Cache.}
    \label{fig:supp_ablation}
\end{figure}

\section{Analysis of the Keyframe Retrieval}
\label{sec:appendix_retrieval}

In this section, we provide a detailed analysis of our keyframe retrieval mechanism for the Shot Cache, addressing its implementation details, hyperparameter sensitivity, and qualitative performance.

\subsection{Qualitative Visualization of the Retrieval Process}
\label{sec:qualitative_viz}
To demonstrate the effectiveness of our framework, Figure~\ref{fig:retrieval} presents a complete case study of our retrieval-augmented generation pipeline. Given a text prompt ("A woman with a blonde ponytail..."), our system searches through a set of candidate keyframes from the shot history. The visualization highlights the Top-3 retrieved frames, which our model has identified as most semantically relevant. For instance, the top-ranked keyframe \#8 provides the core context of the helicopter and water, while others (\#7, \#1) supply more details. This effective retrieval also validates our design choice of setting the Shot Cache size to 3, proving it is adequate for complex, multi-shot contexts.

In addition, the final generated Shot shows how our model synthesizes the visual information from these retrieved keyframes to create a new, coherent scene that accurately matches the prompt. This confirms our framework's ability not only to retrieve relevant concepts but also to use them to generate consistent new content.
\begin{figure}
    \centering
    \includegraphics[width=\linewidth]{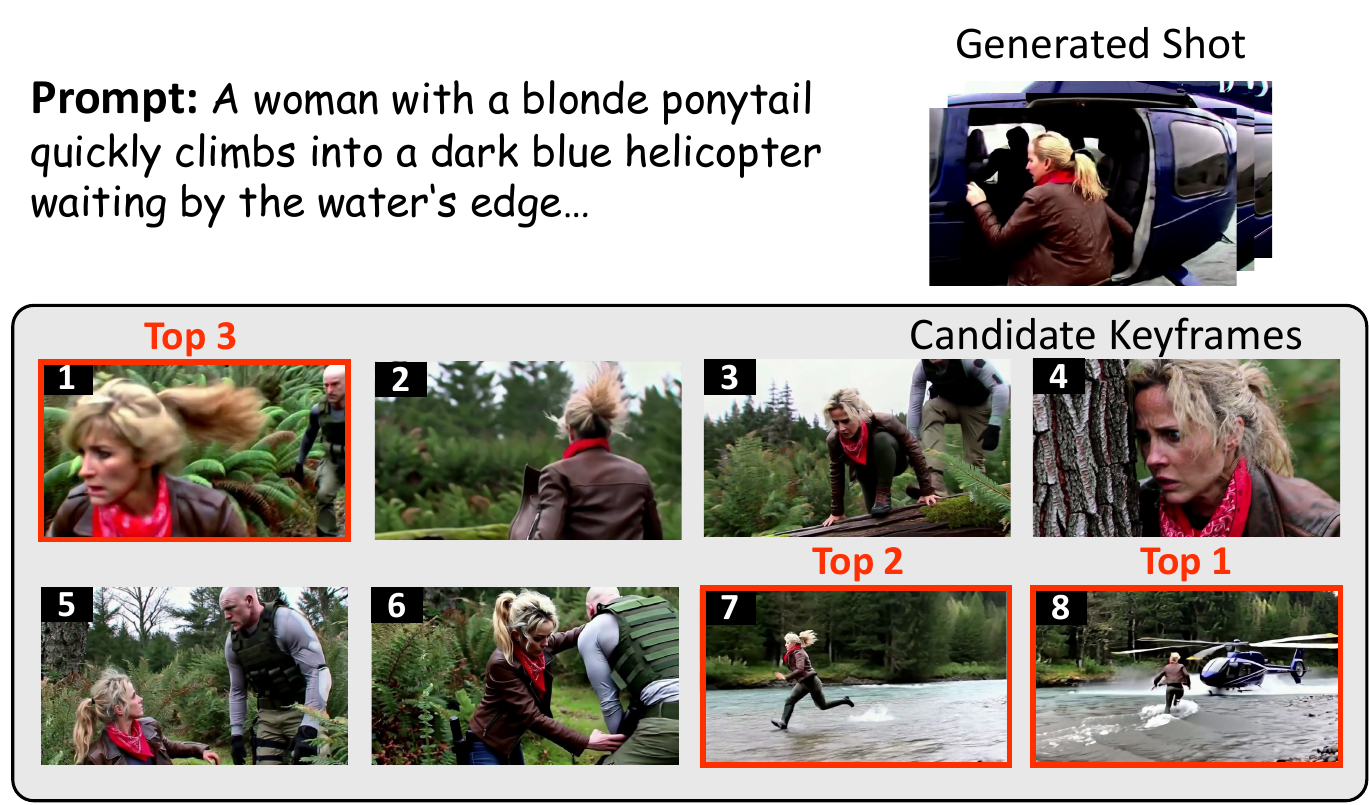}
    \caption{A qualitative example of our prompt-based retrieval. The system correctly identifies the Top-3 keyframes (8, 7, and 1) from the shot history that are most relevant to the text prompt, demonstrating the effectiveness of our approach.}
    \label{fig:retrieval}
\end{figure}

\begin{figure*}[h]
    \centering
    \includegraphics[width=0.95\linewidth]{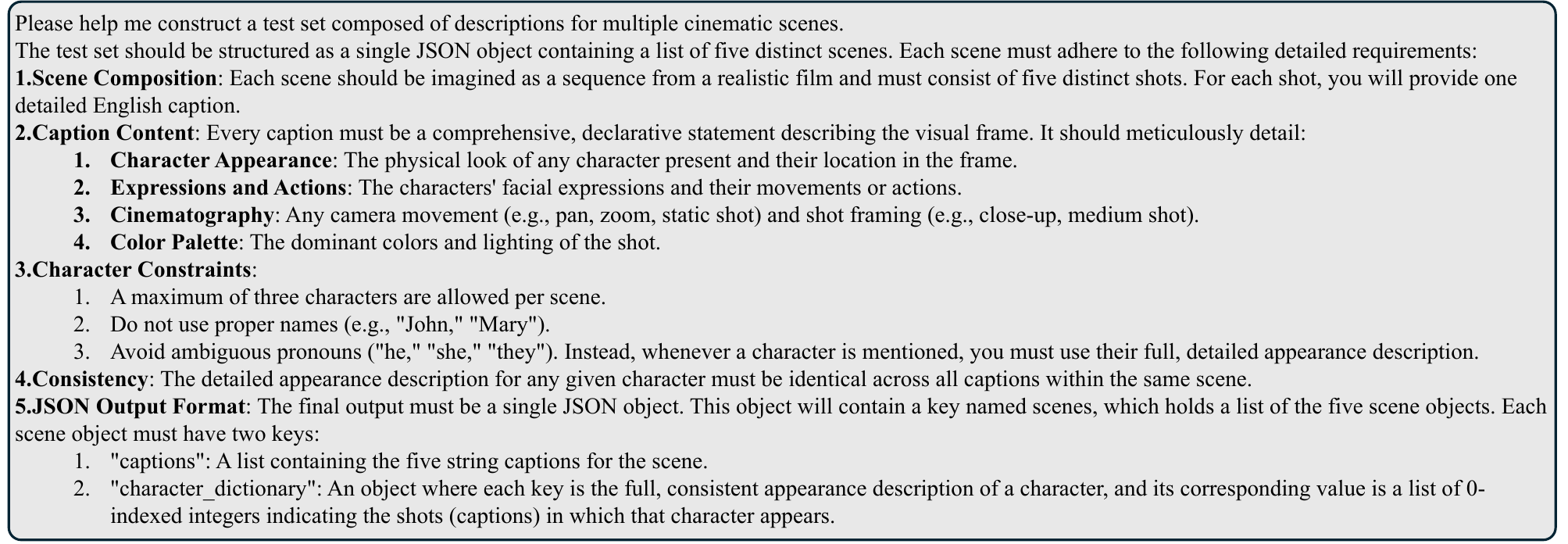}
    \caption{The prompt for test set construction.}
    \label{fig:testset}
\end{figure*}

\subsection{Fault Tolerance via Negative Sampling}
\label{subsec:fault_tolerance}

A key strength of our framework is its robustness to imperfect keyframe retrieval, a resilience cultivated by the negative sampling strategy employed during training. As detailed in the main paper, we randomly introduce irrelevant keyframes into the Shot Cache during the training process. This forces the model to develop a sophisticated, prompt-guided discriminative ability, rather than naively copying all visual information from the cache.

The model learns to cross-reference the visual context from the cache with the semantic guidance from the text prompt. Consequently, it can identify and selectively utilize only the concepts within a reference frame that are relevant to the prompt. As demonstrated in Figure~\ref{fig:supp_negative}, even when the retrieval mechanism erroneously provides an irrelevant keyframe, the model can effectively ignore the distracting visual information. This prevents the injection of incorrect characters or backgrounds, thus safeguarding the narrative's consistency and demonstrating the fault tolerance of our approach.

\begin{figure}
    \centering
    \includegraphics[width=\linewidth]{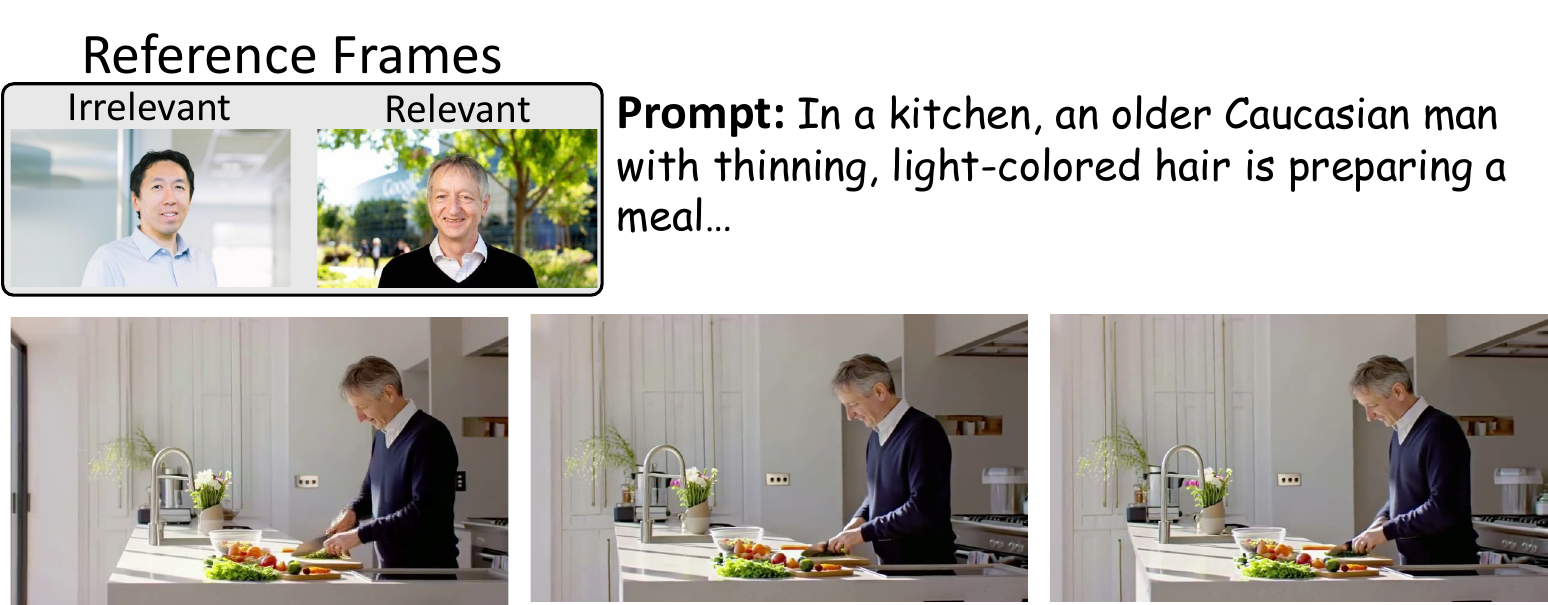}
    \caption{Demonstration of fault tolerance. Despite an irrelevant reference frame in the Shot Cache (left),  our model ignores the incorrect visual information and generates an output that aligns with the text prompt.}
    \label{fig:supp_negative}
\end{figure}

\section{Details of Experiment Settings}
\label{sec:experiment_details}

\subsection{Test Set Construction}
\label{subsec:test_set}
Due to the lack of a suitable public benchmark for text-to-multi-shot video generation, we constructed a new test set to rigorously evaluate our method. To this end, we utilize Gemini 2.5 Pro to generate 20 distinct narrative scenes. Each scene is composed of a sequence of 5 interconnected shots. The prompt for test set construction is shown in Figure~\ref{fig:testset}

\subsection{Evaluation Metrics}
To rigorously quantify the performance of our framework, we evaluate it across three key dimensions: Visual Quality, Consistency, and Text Alignment, as reported in Table 1.

For Visual Quality, we adopt the standardized evaluation protocol from MovieBench, reporting the average per-frame Aesthetics Score (Aes.) and the Inception Score (Incep.).
For Consistency, we assess both character-specific and overall visual style. Character Consistency (Char. Cons.) is calculated by first using Gemini 2.5 Pro with character captions to generate bounding boxes. We then compute the average pairwise CLIP similarity among all cropped images of the same character across different shots within a scene. Overall Consistency (All. Cons.) measures the coherence of the visual style (e.g., background and lighting) by averaging the pairwise CLIP similarity between the keyframes of all shots in the same scene.
For Text Alignment, we again evaluate at both character and overall levels. Character Text Alignment (Char. Align.) is the CLIP similarity between the cropped character images and their corresponding textual descriptions from the prompt. Overall Text Alignment (All. Align.) is calculated as the CLIP similarity between the entire keyframe and its full prompt, averaged across all generated shots.

\section{Computational Efficiency}
Our framework is designed for high computational efficiency. The core of this efficiency lies in our next-chunk prediction paradigm, which contrasts sharply with traditional models that generate content autoregressively based on the full video history.
By conditioning generation on a fixed-size context from our dual-level cache, we avoid the scaling problem where memory requirements grow linearly with video length. This architectural choice yields two primary benefits. First, during training, the constant and modest VRAM footprint allows for larger batch sizes. Second, during inference, the memory consumption remains stable, enabling the generation of arbitrarily long video sequences without the risk of out-of-memory errors. This makes our approach not only effective but also highly scalable and practical for real-world deployment.

Furthermore, our chunk-based approach also reduces the computational cost of the attention mechanism, which is often the primary bottleneck. The complexity of the attention operation is quadratic with respect to the sequence length ($O(n^2)$). To illustrate, consider the task of generating 21 latents. A baseline model generating the full sequence at once, even when utilizing a Shot Cache of 3 keyframes, would process a total sequence length of $n=24$ (3 cache + 21 generated). Its computational cost is therefore proportional to $24^2 = 576$.
In contrast, our method generates the 21 latents in approximately 3.5 times (21 latents / 6 latents per chunk). In each step, the attention is computed over a much shorter, fixed-size context of roughly 11 latents (3 shot cache + temporal cache + chunk). The total computational cost is therefore proportional to approximately $3.5 \times 11^2 = 423.5$. This comparison reveals a substantial reduction in the computational demands of the attention mechanism, leading to less computational budgets in addition to the memory savings.

\end{document}